\documentclass{article}

\usepackage{PRIMEarxiv}

\usepackage[utf8]{inputenc} 
\usepackage[T1]{fontenc}    
\usepackage{hyperref}       
\usepackage{url}            
\usepackage{booktabs}       
\usepackage{amsfonts}       
\usepackage{nicefrac}       
\usepackage{microtype}      
\usepackage{lipsum}
\usepackage{fancyhdr}       
\usepackage{graphicx}       
\graphicspath{{media/}}     

\usepackage{subcaption}
\usepackage{algorithm}
\usepackage{algpseudocode}
\usepackage{amsfonts}
\usepackage{multirow}
\usepackage{array}
\usepackage{dsfont}
\usepackage{amsmath}
\usepackage{amssymb}
\usepackage{caption}

\newtheorem{theorem}{Theorem}
\newtheorem{lemma}{Lemma}

\newtheorem{claim}{Claim}

\newtheorem{assumption}{Assumption}

\title{One-Shot Federated Unsupervised Domain Adaptation with Scaled Entropy Attention and Multi-Source Smoothed Pseudo Labeling}

\author{
  Ali Abedi, Q. M. Jonathan Wu, Ning Zhang \\
  University of Windsor \\
  \texttt{\{abedi3, jwu, ning.zhang\}@uwindsor.ca} \\
   \And
  Farhad Pourpanah \\
  Queen's University \\
  \texttt{farhad.08@gmail.com} \\
}

\begin{document}
\maketitle

\begin{abstract}
Federated Learning (FL) is a promising approach for privacy-preserving collaborative learning. However, it faces significant challenges when dealing with domain shifts, especially when each client has access only to its source data and cannot share it during target domain adaptation. Moreover, FL methods often require high communication overhead due to multiple rounds of model updates between clients and the server. We propose a one-shot Federated Unsupervised Domain Adaptation (FUDA) method to address these limitations. Specifically, we introduce \textit{Scaled Entropy Attention (SEA)} for model aggregation and \textit{Multi-Source Pseudo Labeling (MSPL)} for target domain adaptation. SEA uses scaled prediction entropy on target domain to assign higher attention to reliable models. This improves the global model quality and ensures balanced weighting of contributions. MSPL distills knowledge from multiple source models to generate pseudo labels and manage noisy labels using smoothed soft-label cross-entropy (SSCE). Our approach outperforms state-of-the-art methods across four standard benchmarks while reducing communication and computation costs, making it highly suitable for real-world applications. The implementation code will be made publicly available upon publication. 
\end{abstract}


\section{Introduction}
Federated learning (FL) \cite{mcmahan_communication-efficient_2017} enables decentralized clients to collaboratively train a global model while maintaining data privacy \cite{zhang_survey_2021, shenaj_federated_2023}. Typically, each client trains a local model on its data, and the server aggregates these models to form the global model \cite{wen_survey_2023}. However, this process requires multiple communication rounds, which introduces significant communication and computation overhead, and increases the risk of communication failures \cite{beitollahi_parametric_2024}. Another challenge in conventional FL is domain shift, where data distributions differ significantly across clients (source domains) \cite{feng_kd3a_2021, yi_model-contrastive_2023}, and the distribution of test data (target domain) may also vary from that of the training data \cite{li_federated_2024}. Such heterogeneity limits the applicability of FL in real-world scenarios where data diversity is common \cite{wang_deep_2018}.

One-shot FL \cite{guha_one-shot_2019, su_one-shot_2023, beitollahi_parametric_2024} offers a streamlined approach to address communication challenges by reducing interactions to a single round between clients and the server. This reduction in communication rounds lowers overhead compared to traditional multi-round FL methods.
This strategy is particularly effective in scenarios where repeated communication rounds are impractical or cost-prohibitive \cite{beitollahi_parametric_2024}.

To address the domain shift challenge, multi-source unsupervised domain adaptation techniques have been developed to transfer knowledge from multiple labeled source domains to an unlabeled target domain \cite{sun_survey_2015, li_multi-source_2022}. Federated unsupervised domain adaptation (FUDA) \cite{li_federated_2024} extends this approach to a federated setup, where data remains decentralized across multiple clients. In FUDA, each client represents a labeled source domain, and together they adapt a model to work on an unlabeled target domain while preserving privacy \cite{yin_universal_2022, wei_multi-source_2024}. 

However, current FUDA methods face several limitations. Many require access to labeled source data during adaptation \cite{kang_communicational_2022, peng_federated_2019, pourpanah_federated_2025} or rely on sharing statistical information \cite{yi_model-contrastive_2023}, which is often impractical, and may compromise privacy. Furthermore, many aggregation methods assign a high weights to clients with larger datasets regardless of their performance \cite{yi_model-contrastive_2023}. This overshadows the contributions from clients with smaller datasets \cite{liu_projected_2021, li_joint_2022, qi_model_2024}.

In this study, we propose a novel approach for FUDA within a one-shot FL setup.
Our approach comprises two key components: \textit{(i)} \textit{\textbf{S}caled \textbf{E}ntropy \textbf{A}ttention (SEA)} for model aggregation, and \textit{(ii)} \textit{\textbf{M}ulti-\textbf{S}ource \textbf{P}seudo \textbf{L}abeling (MSPL)} for target domain adaptation. SEA assigns attention weights to source models based on their entropy on the unlabeled target domain. It computes a weighted average of the source models' trained parameters, which forms the global model with higher contributions from more confident models. Then, MSPL generates pseudo labels from multiple source models to distill knowledge into the global model.  
This process adapts the global model to the unlabeled target domain. It also allows the model to benefit from diverse data distributions without compromising privacy.
Unlike existing methods that use thresholding techniques to mitigate noise in pseudo labels \cite{saito_asymmetric_2017, zou_unsupervised_2018}, we adopt smoothed soft-label cross-entropy (SSCE) \cite{szegedy_rethinking_2016, soltany_federated_2025} to manage noise without discarding any labels.
We validate our method on four standard benchmarks: OfficeHome \cite{venkateswara_deep_2017}, Office-31 \cite{hutchison_adapting_2010}, Office-Caltech \cite{gong_geodesic_2012}, and DomainNet \cite{peng_moment_2019}. The results demonstrate the effectiveness of our approach compared to various methods. Additionally, we conducted an ablation study and sensitivity analysis to assess the contribution of different components and the impact of parameters in our proposed approach.

The main contributions of this paper include: \textit{(i)} We propose a novel FUDA framework within a one-shot FL setup. Our approach comprises SEA, a new model aggregation method that assigns attention weights based on scaled prediction entropy, and MSPL, a technique that adapts the global model to an unlabeled target domain using pseudo labels from multiple source models using SSCE loss, \textit{(ii)} We theoretically prove SEA's effectiveness by bounding the aggregated target risk and showing how entropy-based weighting and scaling reduce model discrepancy. Additionally, we establish SSCE’s regularization effect in mitigating noise and improving generalization, and \textit{(iii)} We validate our method with extensive experiments and ablation studies across four benchmark datasets.
\section{Related Works}
\label{sec:related_works}
\textbf{Federated learning.}
FL has gained significant attention as an alternative solution to privacy concerns in distributed machine learning \cite{zhang_survey_2021}.
FedAvg~\cite{mcmahan_communication-efficient_2017} is a widely adapted approach that facilitates decentralized training by enabling clients to perform local model updates. These updates are subsequently aggregated through a model averaging process, where each model is weighted according to the size of training data at the respective client. Building on this, Yurochkin \textit{et al.} \cite{yurochkin_bayesian_2019} introduce Bayesian methods for parameter matching and merging in decentralized settings. Despite these advancements, minimizing communication overhead remains a key challenge in FL. This drives ongoing efforts towards the development of one-shot FL techniques. MA-Echo~\cite{su_one-shot_2023} aligns local models toward a common low-loss region in the parameter space, enabling efficient aggregation with minimal central coordination. FedPFT~\cite{beitollahi_parametric_2024} uses foundation models to transfer Gaussian mixtures of class-conditional features to enable data-free knowledge transfer to improve communication efficiency while maintaining model accuracy.

\noindent \textbf{Multi-source unsupervised domain adaptation.} This category of problems involves transferring knowledge from multiple labeled source domains to an unlabeled target domain to address domain shift. For example, SHOT \cite{liang_we_2020} uses information maximization and self-supervised pseudo-labeling. ABMSDA \cite{zuo_attention-based_2021} minimizes negative transfer effects by focusing on source domains similar to the target using attention mechanisms.

UMAN \cite{yin_universal_2022} uses a pseudo-margin vector for reliable common class identification and aligns source and target distributions without added complexity. SImpAI \cite{venkat_your_2020} leverages classifier agreement to align latent features across source and target domains. CSR \cite{zhou_cycle_2024} leverages source-specific networks and a domain-ensemble network that captures dominant knowledge from each source. 
DECISION \cite{ahmed_unsupervised_2021} adapts multiple source models to the target domain using weighted pseudo-labeling. EAAF \cite{pei_evidential_2024} and Agile MSFDA \cite{li_agile_2024} frameworks further improve adaptation through ensemble strategies and efficient model aggregation. However, these approaches face limitations in federated environments \cite{zhang_survey_2021}, where source domains are trained separately on clients that cannot share data with each other or with the server.

\noindent \textbf{Federated unsupervised domain adaptation.} FUDA tackles the domain shift problem between multiple labeled source domains and an unlabeled target domain while preserving privacy. FADA \cite{peng_federated_2019} addresses domain shift across decentralized devices in a federated environment by extending adversarial domain adaptation with a dynamic attention mechanism. COPA \cite{wu_collaborative_2021} optimizes clients with domain-invariant feature extractors and aggregates them without sharing data across domains. KD3A \cite{feng_kd3a_2021} leverages consensus knowledge and reweighting strategies to mitigate negative transfer and improve adaptation performance. MCKD \cite{niu_mckd_2023} presents a data-free knowledge distillation framework that aligns clients through mutual knowledge sharing. FdFTL \cite{li_federated_2024-1} proposes a federated fuzzy learning method that tackles domain and category shifts without sharing source data. UFDA \cite{liu_ufda_2024} minimizes assumptions in federated domain adaptation by allowing label set diversity among source domains and keeping the target domain’s label set unknown.

\noindent \textbf{Domain adaptation using pseudo labels.} Pseudo-labeling is a learning technique where a model generates artificial labels, or 'pseudo-labels,' for unlabeled data. This approach serves as a crucial mechanism to enhance the adaptation process \cite{li_pseudo_2023}. Wang \textit{et al.} \cite{wang_unsupervised_2020} improve pseudo-label accuracy by introducing selective pseudo-labeling combined with structured prediction, where target samples are clustered to generate more reliable labels. Litrico \textit{et al.} \cite{litrico_guiding_2023} refine pseudo-labels by estimating uncertainty, which allows the classification loss to be reweighted, leading to enhanced accuracy and improved domain adaptation.

Multi-classifier approaches, unlike single-source methods, train several networks to produce more accurate pseudo-labels for unlabeled data \cite{li_pseudo_2023}. Matsuzaki \textit{et al.} \cite{matsuzaki_multi-source_2023} develop a method for domain-adaptive training using multiple source datasets, where soft pseudo-labels are generated by integrating object probabilities from source models and weighted based on the similarity between source domains and the target dataset. SImpAl \cite{venkat_your_2020} assigns pseudo-labels to target samples based on classifier agreement, and only samples where all classifiers predict the same class label are selected. Zheng \textit{et al.} \cite{zheng_rectifying_2021} propose a method to rectify pseudo-label learning by incorporating prediction uncertainty into the model, where prediction variance is used to adjust confidence thresholds dynamically.

\section{Methodology}
\label{sec:methodology}
\subsection{Problem Formulation}

Assume there are $M$ clients, each has access to its local data from a distinct source domain $\mathcal{D}_{i}^{S} = \{x_{ij}, y_{ij}\}_{j=1}^{N_i}$, where $x_{ij} \in \mathcal{X}_i$ and $y_{ij} \in \mathcal{Y}_i$ represent the $j^{th}$ input and its corresponding label of the $i^{th}$ client, and $N_i$ indicates the number of samples for the $i^{th}$ client ($i = 1, 2, \ldots, M$).
Due to privacy constraints, clients cannot share their local data, features, or any statistical information with other clients or the server.
The goal of FUDA is to train a global model $f_\text{g}$ by aggregating $M$ clients, as: 
\begin{equation}
f_{\text{g}} = \sum_{i=1}^{M} w_{i}f_i,
\label{eq:agg_model}
\end{equation}
where $w_i$ is the weight of $i^{th}$ client, and $f_i$ is the $i^{th}$ client trained by minimizing the following objective:
\begin{equation}
\min \mathcal{L}_i(f_i) = \frac{1}{N_i} \sum_{j=1}^{N_i} \ell(f_i(x_{ij}), y_{ij}),
\end{equation}
where $\ell$ represents the cross-entropy loss.

To further adapt the global model to the target domain ${D}^T = \{x_j^T\}_{j=1}^{N_T}$, where $N_T$ is the number of samples in the target domain, we optimize the following objective: 
\begin{equation}
\min \mathcal{L}_{\text{target}}(f_{\text{g}}) = \frac{1}{N_T} \sum_{j=1}^{N_T} \ell(f_{\text{g}}(x_j^T), \mathcal{D}^T).
\end{equation}

\subsection{Model Overview}
\begin{figure*}[th]
\centering
\includegraphics[width=\textwidth]{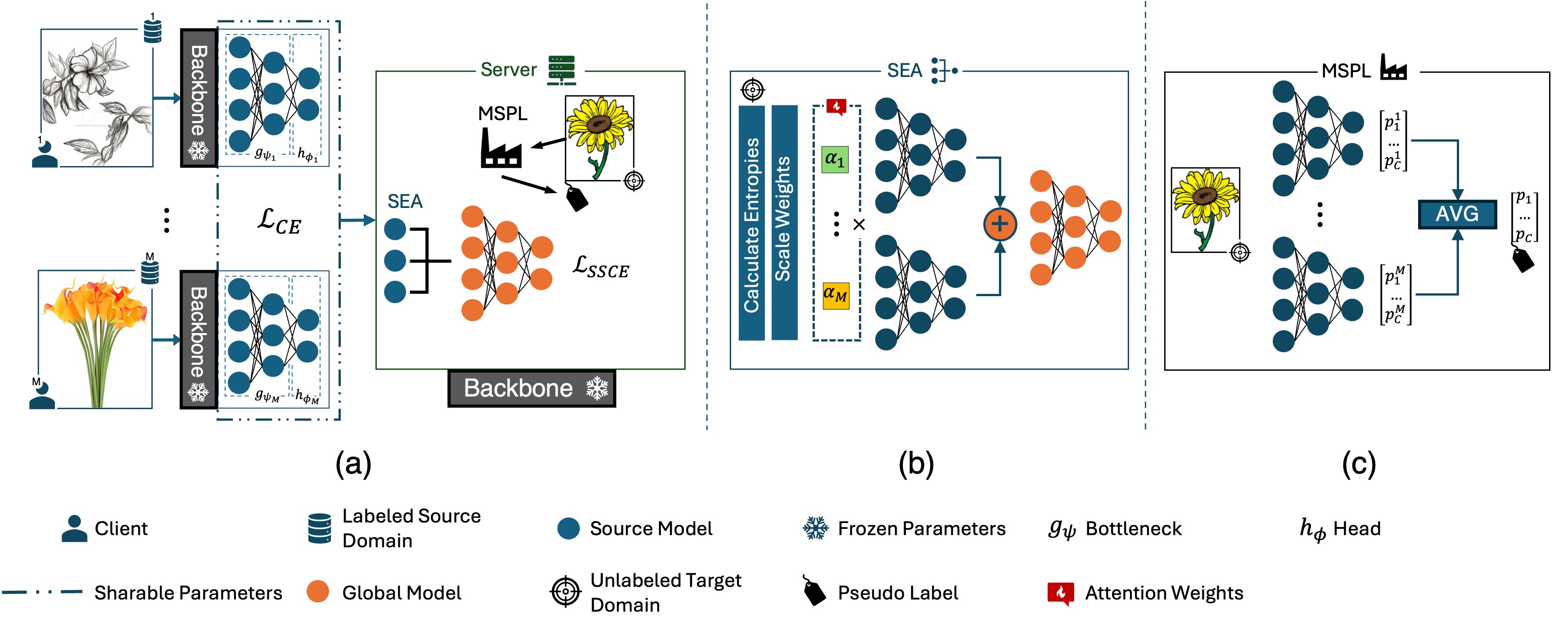}
\caption[Caption for LOF]{Overview of our proposed FUDA framework. \textbf{(a)} Each client trains a local source model and sends only the bottleneck and head parameters to the server, which then aggregates these models with SEA and refines the global model for the target domain using MSPL. \textbf{(b)} SEA aggregation process. It calculates entropy attention weights to generate the global model. \textbf{(c)} MSPL module. It generates pseudo labels for the target domain using source models and trains the global model with SSCE loss.}
\label{fig:model}
\end{figure*}

Figure~\ref{fig:model} shows an overview of our proposed FUDA model, which adopts a one-shot federated learning approach. Each client locally trains a model on its source-domain data and transmits the trained parameters to the server in a single communication round. The server aggregates these parameters using our novel SEA technique to further apply target-domain adaptation. Then we introduce the MSPL module, which generates pseudo labels from multiple source models and refines the global model using SSCE. Our approach preserves data privacy by not sharing raw data or statistical details and significantly reduces communication overhead.

\subsection{Model Architecture}
Each client has a backbone, a bottleneck, and a head layer. The backbone is a frozen pre-trained DINOv2 model \cite{oquab_dinov2_2024}, which is a self-supervised ViT-based model \cite{dosovitskiy_image_2021} known for extracting robust features from unlabeled data. The bottleneck comprises multiple fully connected layers that refine features, with adjustable complexity \cite{abedi_euda_2024}. The head, a single fully connected layer, generates the final predictions.
Clients only train the bottleneck and head layers locally using their respective source domain data. Afterward, each client sends the trained bottleneck and head to the server for aggregation. By freezing the backbone, communication and computational costs are minimized, as only the lightweight components (bottleneck and head) are trained and shared. The overall model for the $i^{th}$ client can be expressed as:
\begin{equation}
  f_i(x_i) = h_{\phi_i}(g_{\psi_i}(b_{\omega}(x_i))),
  \label{eq:overal_arch}
\end{equation}
where $\omega$ represents the frozen backbone weights, and $\psi_i$ and $\phi_i$ represent the trainable weights of the bottleneck and head for the $i^{th}$ client, respectively.

 \subsection{Scaled Entropy Attention}
To aggregate the clients at the server level, SEA first computes the prediction entropy of each client on the unlabeled target domain as: 
\begin{equation}
    \label{eq:entropy}
    H_i(p_i(x)) = - \sum_{k=1}^{C} p_i(x)_k \log(p_i(x)_k),
\end{equation}
where $p_i(x)$ is the probability distribution of the predictions made by the $i^{th}$ clients after applying the softmax function.

Next, the average entropy for each client is calculated over the entire target dataset as:
\begin{equation}
    \label{eq:avg_entropy}
    \bar{H}_i = \frac{1}{N_T} \sum_{j=1}^{N_T} H_i(p_i(x_j)).
\end{equation} 
where $N_T$ is the total number of samples in the target domain.

Finally, each client's entropy is used to calculate attention weights for aggregating the global model:
\begin{equation}
    \label{eq:entropy_weights}
    w_i' = \frac{1}{\bar{H}_i},
\end{equation}
and the normalized weights are given by:
\begin{equation}
    \label{eq:normalized_weights}
    w_i = \frac{w_i'}{\sum_{j=1}^{M} w_j'} = \frac{\frac{1}{\bar{H}i}}{\sum_{j=1}^{M} \frac{1}{\bar{H}_j}}.
\end{equation}

To prevent uniform weight distributions when entropy values cluster closely, we introduce scaling based on the deviation of weights from the mean:
\begin{equation}
\label{eq:scaling}
        w_i = \left( \frac{w'_i}{\mu_{w'}} \right)^2,
\end{equation}
where $\mu_{w'}$ is the mean entropy weights, given by:
\begin{equation}
\mu_{w'} = \frac{1}{M} \sum_{j=1}^{M} w'_j.
\end{equation}

Without scaling, lower-confidence models exert a similar influence to high-confidence models when entropy values are near average. This reduces the effectiveness aggregation on the unlabeled target domain, which may lead to suboptimal global model initialization. Our proposed scaling method increases the weights of models significantly deviating from the average and emphasizes confident models while ensuring meaningful contributions from all. The pseudocode of the SEA aggregation is shown in Algorithm \ref{alg:seba}. Section \ref{secsec:sea} provides a theoretical justification for SEA.
\begin{center}
\scalebox{1}{
\begin{minipage}{0.70\textwidth}
\begin{algorithm}[H]
\caption{SEA Aggregation.}
\label{alg:seba}
\begin{algorithmic}[1]
\State \textbf{Input:} Pre-trained source models $\{f_1, f_2, ..., f_M\}$, target domain data $\mathcal{D}_T$
\State \textbf{Output:} Global model $f_{\text{g}}$

\State \textbf{Step 1: Calculate Entropy for Each Model}
\For{each model $f_i$}
    \State Compute $\bar{H}_{i}$ for $f_i$ on the $\mathcal{D}_T$
\EndFor

\State \textbf{Step 2: Calculate Attention Weights}
\For{each model $f_i$}
    \State Compute the $w'_i$ for $f_i$
\EndFor
\State Compute the $\mu_{w'}$ across all models
\For{each model $f_i$}
    \State  $w_i = \left( \frac{w'_i}{\mu_{w'}} \right)^2$
\EndFor
\State Normalize weights

\State \textbf{Step 3: Aggregate Models}
\State Generate global model $f_{\text{g}} = \sum_{i=1}^{M} w_{i}f_i$
\end{algorithmic}
\end{algorithm}
\end{minipage}
}
\end{center}

\subsection{Multi-Source Pseudo Labeling}
\label{sec:mspl}
MSPL aims to further adapt the global model to the target domain. To achieve this, MSPL enables each local model to generate prediction (logit) for each unlabeled sample in the target domain. Then, the average of the logits across all source models for each target sample $x_j^T$ is computed to assign a pseudo label $\tilde{y}_j^T$ to the $j^{th}$ sample, as:
\begin{equation}
    \tilde{y}_j^T = \frac{1}{M} \sum_{i=1}^{M} f_i(x_j^T).
\end{equation}
Once the pseudo labels are generated, they are used to adapt the global model $f_{\text{g}}$ to the unlabeled target domain.
\begin{assumption}
\label{assumption_3}
Pseudo labels \(\hat{y}\) estimate the true labels \( y \), but due to domain shift and model uncertainty, they contain noise. We define the noisy pseudo label as:

\begin{equation}
    \label{eq:noise}
    \tilde{y}_j^T = y_j + \eta_j,
\end{equation}
where $y_j$ is the true, but unknown, label for the target domain and \( \eta \) represents the pseudo-label noise.
\end{assumption}

\begin{lemma}
\label{lemma_ce}
Standard cross-entropy encourages the model to place nearly all probability mass on a single class. This leads to overconfidence and reduced generalization.

The proof is provided in Supplementary Materials (Section \ref{proof:lemma_ce}).
\end{lemma}

\begin{claim}
\label{claim_ls}
To mitigate noise, we apply label smoothing \cite{szegedy_rethinking_2016}, which replaces each pseudo label with a weighted mixture of itself and a uniform distribution:
\end{claim}

\begin{equation}
    \tilde{y}' = (1 - \epsilon) \tilde{y} + \frac{\epsilon}{C}.
\end{equation}

\begin{theorem}
\label{theorem_ls} Applying label smoothing in Claim \ref{claim_ls} to cross-entropy loss regularizes the model by preventing excessive confidence in a single class, thereby improving generalization and reducing overfitting to noisy labels. The smoothing factor $\epsilon$ controls confidence levels, ensuring the model remains uncertain when necessary. This mitigates the effect of the noise term $\eta_j$, making pseudo labels more reliable and enhancing overall generalization. 

Proof for this theorem is provided in Supplementary Materials, Section \ref{proof:theorem_ls}.
\end{theorem}

By applying label smoothing, the SSCE loss for the $j^{th}$ sample $x_j^T$ of the target domain is given by:
\begin{equation}
\label{eq:ssce}
    \mathcal{L}_{\text{smooth}}(x_j^T) =
    - \sum_{c=1}^{C} \left[ (1 - \epsilon) \cdot \tilde{y}_j^T(c) + \frac{\epsilon}{C} \right] \log(p_{\text{g}}(x_j^T)_c),
\end{equation}
where $p_{\text{g}}(x_j^T)_c$ is the predicted probability for class $c$ from the global model. Consequently, the final loss for the target domain training is computed by averaging the loss over all the target samples.
\begin{equation}
    \mathcal{L}_{SSCE} = \frac{1}{N_T} \sum_{j=1}^{N_T} \mathcal{L}_{\text{smooth}}(x_j^T), 
\end{equation}
where  $N_T$ is the total number of target domain samples.
By generating pseudo labels through the MSPL module and applying the SSCE loss, the model can effectively fine-tune the global model to perform better on the target domain, despite the lack of true labels. 

\subsection{Theoretical Justification of SEA}
\label{secsec:sea}

Intuitively, SEA queries each model: 'How certain are you about your predictions on the target data?' Each model responds with an entropy value, where lower entropy indicates higher confidence. By assigning greater weights to more confident models, we ensure a more generalized global model adapted to the target domain \cite{louizos_multiplicative_2017, sensoy_evidential_2018}.
Assume the target risk of a classifier $f$ is defined as:

\begin{equation}
R_T(f) = \mathbb{E}_{(x,y)\sim\mathcal{D}^T} \left[ \ell\bigl(f(x), y\bigr) \right],
\end{equation}
where $\ell(\cdot,\cdot)$ is a suitable loss function, e.g., cross-entropy.

\begin{assumption}
\label{assumption_1}
Each source classifier $f_i$ is well-calibrated on the target domain $\mathcal{D}^T$ so that its true target risk $R_T(f_i)$ is approximately proportional to its average predictive uncertainty, measured by the average entropy $\bar{H}_i$. That is,

\begin{equation}
R_T(f_i) \approx k\, \bar{H}_i, \quad \text{where } \exists k > 0.
\end{equation}
\end{assumption}

\begin{lemma}
\label{lemma_1} Under Assumption \ref{assumption_1}, if for two classifiers $f_i$ and $f_j$ we have:
\begin{equation}
    \bar{H}_i < \bar{H}_j,    
\end{equation}
then:
\begin{equation}
    R_T(f_i) < R_T(f_j).
\end{equation}
\textbf{Proof.} Based on Assumption \ref{assumption_1}, if $\bar{H}_i < \bar{H}_j$, then
\begin{equation}
    k\,\bar{H}_i < k\,\bar{H}_j,
\end{equation}
implying $R_T(f_i) < R_T(f_j)$.
\hfill \(\square\)
\end{lemma}

\begin{claim}
\label{claim_1}
If the optimal weighting to minimize the aggregated target risk were to weight each source classifier inversely proportional to its target risk, then the ideal weight would be:

\begin{equation}
    w_i^\ast \propto \frac{1}{R_T(f_i)},
\end{equation}
and under Assumption \ref{assumption_1}, we have:
\begin{equation}
    w_i^\ast \propto \frac{1}{\bar{H}_i}.
\end{equation}

wBased on this, we define the entropy weights as shown in Eq. \ref{eq:entropy_weights}, and the normalized weights are given by Eq. \ref{eq:normalized_weights}. This choice favors models with lower average entropy, which (by Lemma \ref{lemma_1}) have lower target risk.
\end{claim}

\begin{theorem}[Bound on Aggregated Target Risk]
\label{theorem_1}
Let the aggregated classifier be defined using Eq. \ref{eq:agg_model}, with weights $w_i$ as defined in Claim \ref{claim_1}. Then, under Assumption \ref{assumption_1} and standard multi-source domain adaptation bounds \cite{mansour_domain_2008}, the target risk of the aggregated classifier satisfies:
\begin{equation}
R_T(f_g) \lesssim k\sum_{i=1}^{M} w_i\, \bar{H}_i + \Phi,
\end{equation}
where $\Phi$ represents additional divergence terms capturing the discrepancy between the source and target distributions (proof is provided in Supplementary Materials (Section \ref{proof:theorem_1})).

Since the weights $w_i$ are chosen to be inversely proportional to $\bar{H}_i$, this weighted sum is minimized compared to uniform weighting, leading to a lower bound on $R_T(f_g)$.
\end{theorem}

However, our experiments (Sections \ref{sec:ablation} and \ref{sec:hp_sen}) show that this naive weight calculation is inadequate. 

\begin{assumption}
\label{assumption_2}
When entropy values are closely clustered, the resulting weight distribution becomes nearly uniform.
\begin{equation}
|\bar{H}_{i} - \bar{H}_{j}| \approx 0 \quad \Rightarrow \quad |w'_i - w'_j| \approx 0.
\end{equation}
\end{assumption}

\begin{lemma}
\label{lemma_2}
Under Assumption \ref{assumption_2}, the entropy weights approach their mean (proof is provided in Supplementary Materials (Section \ref{proof:lemma_2})):

\begin{equation}
w'_i \approx \mu_{w'}.
\end{equation}

By Assumption \ref{assumption_2}, the normalized entropy weights converge toward their mean. This results in a nearly uniform weight distribution. This uniformity, under Theorem \ref{theorem_1}, reduces the efectiveness of aggregation process with respect to the unabeled target domain, as lower-confidence models receive similar influence as high-confidence ones.
\end{lemma}

\begin{claim}
\label{claim_2}
By Assumption \ref{assumption_2} and Lemma \ref{lemma_2}, the entropy weights converge toward their mean, leading to a nearly uniform weight distribution that diminishes differentiation among classifiers. To counteract this effect and ensure models with lower entropy retain greater influence, we introduce a scaling transformation as shown in Eq. \ref{eq:scaling}.
\end{claim}

This balances weighting by amplifying confident models while preventing the complete downweighting of others. Notably, larger deviations from the mean result in stronger scaling.
Dividing each value by the mean $\mu_{w'}$ in Eq. \ref{eq:scaling} yields values greater than 1 for entropies exceeding the mean, and vice versa. This scaling operation amplifies the differences: values greater than 1 are further amplified, while those less than 1 are reduced. 

Finally, the $f_\text{g}$ is calculated using Eq. \ref{eq:agg_model}. 

\section{Experiments}
\label{sec:experiments}

\subsection{Datasets} 
\begin{figure*}[htbp]
\centering
\includegraphics[width=\textwidth]{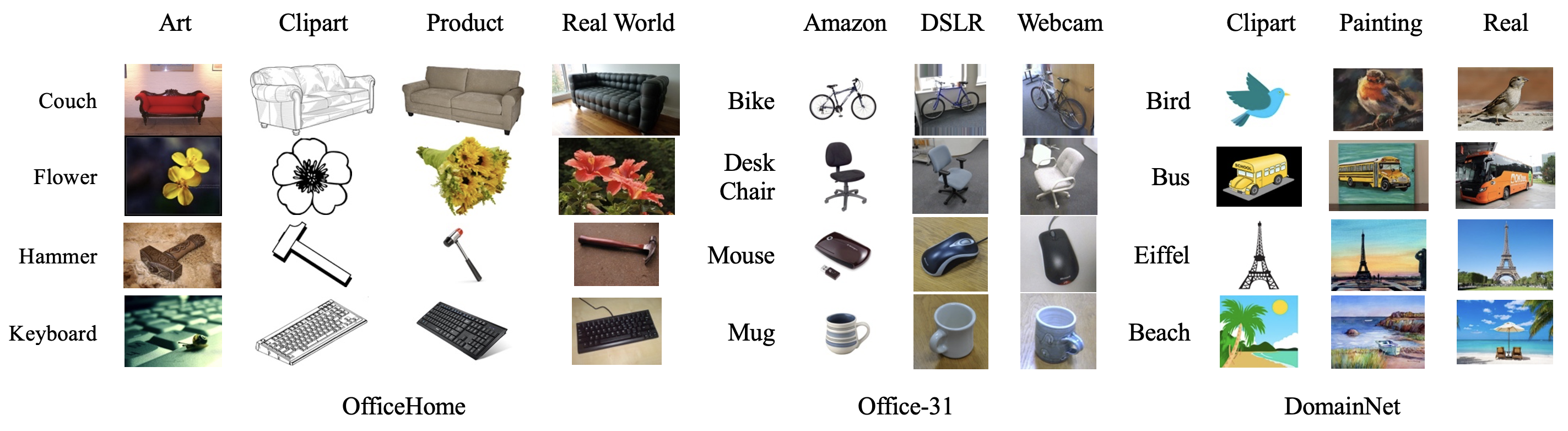}
\caption{Samples from three different datasets, showcasing various domains and classes from OfficeHome, Office-31, and DomainNet}
\label{fig:dataset_samples}
\end{figure*}

We assess the effectiveness of our proposed method using four standard benchmarks. \textbf{OfficeHome} \cite{venkateswara_deep_2017}, containing 15,500 images from four domains, which are Art, Clipart, Product, and Real World, across 65 categories; \textbf{Office-31} \cite{hutchison_adapting_2010}, which contains 4,652 images from three domains namely Amazon, Webcam, and DSLR, across 31 categories; \textbf{Office-Caltech} \cite{gong_geodesic_2012}, with 2,533 images from four domains, Amazon, Webcam, DSLR, and Caltech, covering 10 categories; and \textbf{DomainNet} \cite{peng_moment_2019}, which features 48,129 images from six domains, Clipart, Real, Sketch, Infograph, Painting, and Quickdraw, spanning 345 categories. Sample images for OfficeHome, Office-31, and DomainNet are shown in Figure \ref{fig:dataset_samples}.

\subsection{Implementation details} 
Following prior FUDA works, we assume that each client has access to one source domain, while the server has access to the unlabeled target domain. Thus, for a dataset with $L$ domains, the number of clients (source models) is $M = L-1$. We use a batch size of 32 for all datasets. Local models on each client are trained for 20 epochs, while the global model with the MSPL module is trained for 10 epochs.  The learning rate is set to 3e-2 with 5\% warmup phase. We use SGD with a momentum of 0.9. For hyperparameters, as discussed in Section \ref{sec:hp_sen}, the smoothing factor $\epsilon$ in Eq. \ref{eq:ssce} is set to 0.9.

Following \cite{abedi_euda_2024}, we use a 4-layer base bottleneck (2048, 1024, 512, and 256 nodes) for OfficeHome, Office-31, and Office-Caltech. For DomainNet, we use a small 256-node bottleneck. This setup produces the best results while minimizing parameters and communication overhead. Also, we have evaluated our proposed methods with various backbones and the results are in Section \ref{exp:backbones}.

\subsection{Baselines} We compare the performance of our method with several SOTA approaches based on the availability of results across different benchmarks.
We report the results of SHOT \cite{liang_we_2020}, SImpAl \cite{venkat_your_2020}, and DECISION \cite{ahmed_unsupervised_2021} across all benchmarks. For Office-Caltech and DomainNet, we include results from FADA \cite{peng_federated_2019} and KD3A \cite{feng_kd3a_2021}, while for OfficeHome, Office-31, and DomainNet, we report results from FdFTL \cite{li_federated_2024-1}, EAAF \cite{pei_evidential_2024}, and CSR \cite{zhou_cycle_2024}. Additionally, the results of UMAN \cite{yin_universal_2022} and UFDA \cite{liu_ufda_2024} are provided for OfficeHome and Office-31. For DomainNet, we report results of ABMSDA \cite{zuo_attention-based_2021} and Bi-ATEN \cite{li_agile_2024}, while MCKD \cite{niu_mckd_2023} and COPA \cite{wu_collaborative_2021} results are included exclusively for Office-Caltech.
Among these methods, SImpAl \cite{venkat_your_2020}, CSR \cite{zhou_cycle_2024}, ABMSDA \cite{zuo_attention-based_2021}, UMAN \cite{yin_universal_2022}, SHOT \cite{liang_we_2020}, EAAF \cite{pei_evidential_2024}, and Bi-ATEN \cite{li_agile_2024} are multi-source unsupervised domain adaptation approaches, while FdFTL \cite{li_federated_2024-1}, UFDA \cite{liu_ufda_2024}, FADA \cite{peng_federated_2019}, COPA \cite{wu_collaborative_2021}, MCKD \cite{niu_mckd_2023}, and KD3A \cite{feng_kd3a_2021} are FUDA techniques.

\begin{table*}[tb]
\caption{The results of methods on OfficeHome and Office-31. `FL' and 'OS' indicate federated learning and one-shot FL, respectively. Bold indicates the best results.}
\label{table:comp_oh}
\centering
\def\arraystretch{1.2}
\resizebox{0.9\textwidth}{!}{
\begin{tabular}{l|cc|cccc|c|ccc|c}
\toprule
\multirow{2}{*}{Model}  & \multirow{2}{*}{FL} & \multirow{2}{*}{OS} & \multicolumn{5}{c|}{OfficeHome} & \multicolumn{4}{c}{Office-31} \\ \cline{4-12}
& & & $\rightarrow$A & $\rightarrow$C & $\rightarrow$P & $\rightarrow$R & Avg. & $\rightarrow$A & $\rightarrow$D & $\rightarrow$W & Avg. \\ \midrule
SImpAl \cite{venkat_your_2020}          & $\times$     & $\times$         & 73.4           & 62.4           & 81.0           & 82.7           & 74.8 & 71.2           & 81.0           & 97.9           & 89.5 \\
UMAN \cite{yin_universal_2022}          & $\times$     & $\times$         & 79.0           & 64.7           & 81.1           & 87.1           & 78.0 & \textbf{90.2}  & 94.5           & 94.5           & 93.1 \\
CSR \cite{zhou_cycle_2024}              & $\times$     & $\times$         & 76.7           & 86.8  & 71.4           & 85.5           & 80.1 & 76.7           & \textbf{100.0} & \textbf{99.6}  & 92.7 \\
SHOT \cite{liang_we_2020}               & $\times$ & $\times$         & 72.1           & 57.2           & 83.0           & 81.5           & 73.5 & 75.0           & 94.9           & 97.8           & 81.5 \\
DECISION \cite{ahmed_unsupervised_2021} & $\times$ & $\times$         & 74.5           & 59.4           & 84.4           & 83.6           & 75.5 & 75.4           & 99.6           & 98.4           & 91.1 \\
EAAF \cite{pei_evidential_2024}         & $\times$ & $\times$         & 75.9           & 61.5           & 86.7           & 84.5           & 77.2 & 76.8           & 99.8           & 98.7           & 91.7 \\
FdFTL \cite{li_federated_2024-1}          & $\checkmark$ & $\times$     & 73.5           & 61.7           & 83.9           & 83.8           & 75.7 & 74.3           & 99.8           & 98.9           & 91.0 \\

UFDA \cite{liu_ufda_2024}          & $\checkmark$     & $\times$         & 80.3           & 62.3           & 82.3           & 88.9           & 78.5 & 80.0  & 96.6        & 96.5           & 91.0 \\
\textbf{SEA (Ours)}                    & $\checkmark$ & $\checkmark$ & 75.9           & 71.3           & 89.1           & 83.6           & 80.0 & 77.6           & 99.8           & 99.2           & 92.2 \\
\textbf{SEA + MSPL (Ours)}             & $\checkmark$ & $\checkmark$ & \textbf{83.2}  & \textbf{76.6}           & \textbf{91.2}  & \textbf{90.8}  & \textbf{85.4} & 82.2           & 99.8           & 99.2           & \textbf{93.7} \\ \bottomrule
\end{tabular}%
}
\end{table*}

\begin{table}[tb]
\caption{The results of methods on Office-Caltech. `FL' and 'OS' indicate federated learning and one-shot FL, respectively. Bold indicates the best results.}
\label{table:comp_oc}
\centering
\def\arraystretch{1.2}
\resizebox{0.65\textwidth}{!}{
\begin{tabular}{l|cc|cccc|c}
\toprule
Model                                   & FL           & OS                   & $\rightarrow$A & $\rightarrow$C & $\rightarrow$D & $\rightarrow$W & Avg. \\ \midrule
SImpAl \cite{venkat_your_2020}          & $\times$     & $\times$         & 95.6           & 94.6           & \textbf{100.0}  & \textbf{100.0} & 97.5 \\

SHOT \cite{liang_we_2020}               & $\times$ & $\times$         & 95.7           & 95.8           & 96.8            & 99.6           & 97.0 \\
DECISION \cite{ahmed_unsupervised_2021} & $\times$ & $\times$         & 95.9           & 95.9           & \textbf{100.0}  & 99.6           & 98.0 \\
FADA \cite{peng_federated_2019}         & $\checkmark$ & $\times$     & 84.2           & 88.7           & 87.1            & 88.1           & 87.1 \\
COPA \cite{wu_collaborative_2021}   & $\checkmark$     & $\times$     & 95.8           & 94.6           & 99.6  & 99.8 & 97.5 \\
MCKD \cite{niu_mckd_2023}   & $\checkmark$     & $\times$     & 95.8  & 94.9           & \textbf{100.0}  & 99.9 & 97.7 \\
KD3A \cite{feng_kd3a_2021}  & $\checkmark$ & $\times$         & \textbf{97.4}  & 96.4           & 98.4  & 99.7           & 97.9 \\
\textbf{SEA (Ours)}                    & $\checkmark$ & $\checkmark$ & 96.1  & \textbf{97.2}  & 98.7            & 99.7           & 97.9 \\
\textbf{SEA + MSPL (Ours)}             & $\checkmark$ & $\checkmark$ & 96.1  & \textbf{97.2}  & 99.4            & 99.7           & \textbf{98.1} \\ \bottomrule
\end{tabular}%
}
\end{table}

\begin{table}[tb]
\caption{The results of methods on DomainNet. `FL' and 'OS' indicate federated learning and one-shot FL, respectively. Bold indicates the best results.}
\label{table:comp_dn}
\centering
\def\arraystretch{1.2}
\resizebox{0.75\textwidth}{!}{
\begin{tabular}{l|cc|cccccc|c}
\toprule
Model                                   & FL           & OS                   & $\rightarrow$C & $\rightarrow$I & $\rightarrow$P & $\rightarrow$Q & $\rightarrow$R & $\rightarrow$S & Avg. \\ \midrule
ABMSDA \cite{zuo_attention-based_2021} & $\times$     & $\times$        & 66.9           & 25.3           & 55.8           & 18.2           & 64.1 & 55.2 & 47.6 \\
SImpAl \cite{venkat_your_2020}          & $\times$     & $\times$         & 66.4           & 26.5           & 56.6           & 18.9           & 68.0 & 55.5 & 48.6 \\ 
CSR \cite{zhou_cycle_2024}              & $\times$     & $\times$         & 73.0           & 28.1           & 58.8           & 26.0           & 71.1 & 60.7 & 52.9 \\

SHOT \cite{liang_we_2020}               & $\times$ & $\times$         & 61.7           & 22.2           & 52.6           & 12.2           & 67.7 & 48.6 & 44.2 \\
DECISION \cite{ahmed_unsupervised_2021} & $\times$ & $\times$         & 61.5           & 21.6           & 54.6           & 18.9           & 67.5 & 51.0 & 45.9 \\ 
EAAF \cite{pei_evidential_2024}         & $\times$ & $\times$         & 68.8           & 26.2           & 58.4           & 23.5           & 70.8 & 57.1 & 50.7 \\
Bi-ATEN \cite{li_agile_2024}        & $\times$ & $\times$         & \textbf{77.0}           & \textbf{38.5}           & \textbf{68.6}           & \textbf{25.0}          & \textbf{83.6} & \textbf{64.9} & \textbf{59.6} \\
FADA \cite{peng_federated_2019}         & $\checkmark$ & $\times$     & 45.3           & 16.3           & 38.9           & 7.9           & 46.7 & 26.8 & 30.3 \\
FdFTL \cite{li_federated_2024-1}          & $\checkmark$ & $\times$     & 70.5           & 25.9           & 58.1           & 21.3           & 70.6 & 59.0 & 50.9 \\
KD3A \cite{feng_kd3a_2021}      & $\checkmark$ & $\times$          & 72.5           & 23.4           & 60.9           & 16.4           & 72.7 & 60.6 & 51.1 \\
\textbf{SEA (Ours)}                    & $\checkmark$ & $\checkmark$ & 66.0           & 29.6           & 59.8           & 8.5           & 73.5 & 58.6 & 49.3 \\
\textbf{SEA + MSPL (Ours)}             & $\checkmark$ & $\checkmark$ & 69.7           & 33.5           & 60.9           & 19.7           & 74.9 & 60.8 & 53.2 \\ \bottomrule
\end{tabular}%
}
\end{table}

\subsection{Results}
\subsubsection{Performance}
Tables \ref{table:comp_oh}, \ref{table:comp_oc}, and \ref{table:comp_dn} show the accuracy rates of our proposed method and the baselines on four benchmark datasets. For our method, we report the accuracy of both SEA aggregation and SEA + MSPL. Overall, the SEA + MSPL model consistently outperforms all baselines in terms of mean accuracy across nearly all datasets. Specifically, for the OfficeHome dataset, SEA + MSPL achieves a 5.3\% improvement over the best SOTA method (Table \ref{table:comp_oh}). Similarly, for the Office-31 and Office-Caltech datasets, SEA + MSPL surpasses all baselines (Tables \ref{table:comp_oh} and Table \ref{table:comp_oc}, respectively). Finally, SEA + MSPL demonstrates performance on par with the baselines for the DomainNet dataset (Table \ref{table:comp_dn}).

Notably, our method obtains these results with a significantly small number of trainable parameters as the backbone is frozen during the training process (see Table \ref{table:tp} in the Supplementary Materials). Furthermore, it operates under highly restrictive constraints, solving the FUDA problem in a one-shot FL setup. This underscores the robustness of our approach, offering strong performance and efficiency while preserving privacy and minimizing communication overhead. Importantly, our method maintains strong performance across different target domains, reflecting its robustness. While certain baselines may achieve higher accuracy on specific target domains, our model consistently outperforms them on average, highlighting its stability and generalization ability.

Table \ref{table:tp} compares the number of trainable parameters in our proposed method with other well-known approaches. As shown, our model requires significantly fewer trainable parameters, which is particularly important in FL, where clients share their trained models with the server. Notably, in our method, the number of trainable parameters remains independent of the backbone, as the backbone is non-trainable, ensuring efficiency regardless of the chosen architecture.

\begin{table}[htb]
\caption{The number of trainable parameters by our method vs. others.}
\label{table:tp}
\centering
\def\arraystretch{1.2}
\resizebox{0.65\textwidth}{!}{
\begin{tabular}{l|cc}
\toprule
Model & \# Trainable Parameters & Backbone
 \\ \midrule
 DECISION \cite{ahmed_unsupervised_2021} & 120.1 M & ResNet-50 \cite{he_deep_2016} \\
 Bi-ATEN \cite{li_agile_2024} & 10.6 M & Swin \cite{liu_swin_2021} \\
 ATEN \cite{li_agile_2024} & 4.9 M & Swin \cite{liu_swin_2021} \\
 \textbf{Ours (Base Bottleneck)} & \textbf{4.4 M} & ViT (DINOv2) \cite{oquab_dinov2_2024} \\
 \textbf{Ours (Small Bottleneck)} & \textbf{0.3 M} & ViT (DINOv2) \cite{oquab_dinov2_2024} \\ \bottomrule
\end{tabular}%
}
\end{table}

\subsubsection{Ablation Study}
\label{sec:ablation}
Here, we conduct an ablation study to evaluate the effectiveness of SEA and MSPL in our proposed model by systematically removing each component. As shown in Table \ref{table:ablation}, first, we replaced $\mathcal{L}_{SSCE}$ with $\mathcal{L}_{CE}$, and trained the model with pseudo labels without soft-label smoothing. This demonstrates that using $\mathcal{L}_{SSCE}$ helps manage noise in pseudo labels, leading to better knowledge distillation from source domains to the global model. Next, we removed MSPL and evaluated SEA without further adaptation to the target domain. The results show that MSPL significantly enhances the accuracy of the global model on the target domain. We then removed the scaling from SEA, aggregating models using simple entropy-based attention weights, which highlights the effectiveness of our proposed scaling method. Finally, we removed the entropy-based attention altogether, using simple averaging for model aggregation. As can be seen in the results, attention weights considerably boost model performance.

\begin{table}[t]
\caption{The ablation study on OfficeHome. Bold indicates the best results.}
\label{table:ablation}
\centering
\def\arraystretch{1.2}
\resizebox{0.7\textwidth}{!}{
\begin{tabular}{l|cccc|c}
\toprule
Method                    & $\rightarrow$A & $\rightarrow$C & $\rightarrow$P & $\rightarrow$R & Avg. \\ \midrule
\textbf{SEA + MSPL} & \textbf{83.2} & 76.6 & \textbf{91.2} & \textbf{90.8} & \textbf{85.4} \\
SEA + MSPL w/o $\mathcal{L}_{SSCE}$ & 77.8 & \textbf{85.4} & 85.4 & 86.3 & 83.7 \\

\textbf{SEA}                       & 75.9           & 71.3           & 89.1           & 83.6           & 80.0 \\
w/ Entropy-based Attention, w/o Scaling                        & 71.6           & 71.5           & 88.8           & 82.4           & 78.6 \\
w/o Entropy-based Attention        & 64.5           & 71.7           & 88.3           & 79.4           & 76.0 \\
\bottomrule
\end{tabular}%
}
\end{table}

\subsubsection{Sensitivity Analysis}
\label{sec:hp_sen}

In this section, we conduct sensitivity analyses to show the impact of different hyperparameters and components on the performance of our proposed method.

\begin{figure*}[htbp]
\centering
\includegraphics[width=0.5\textwidth]{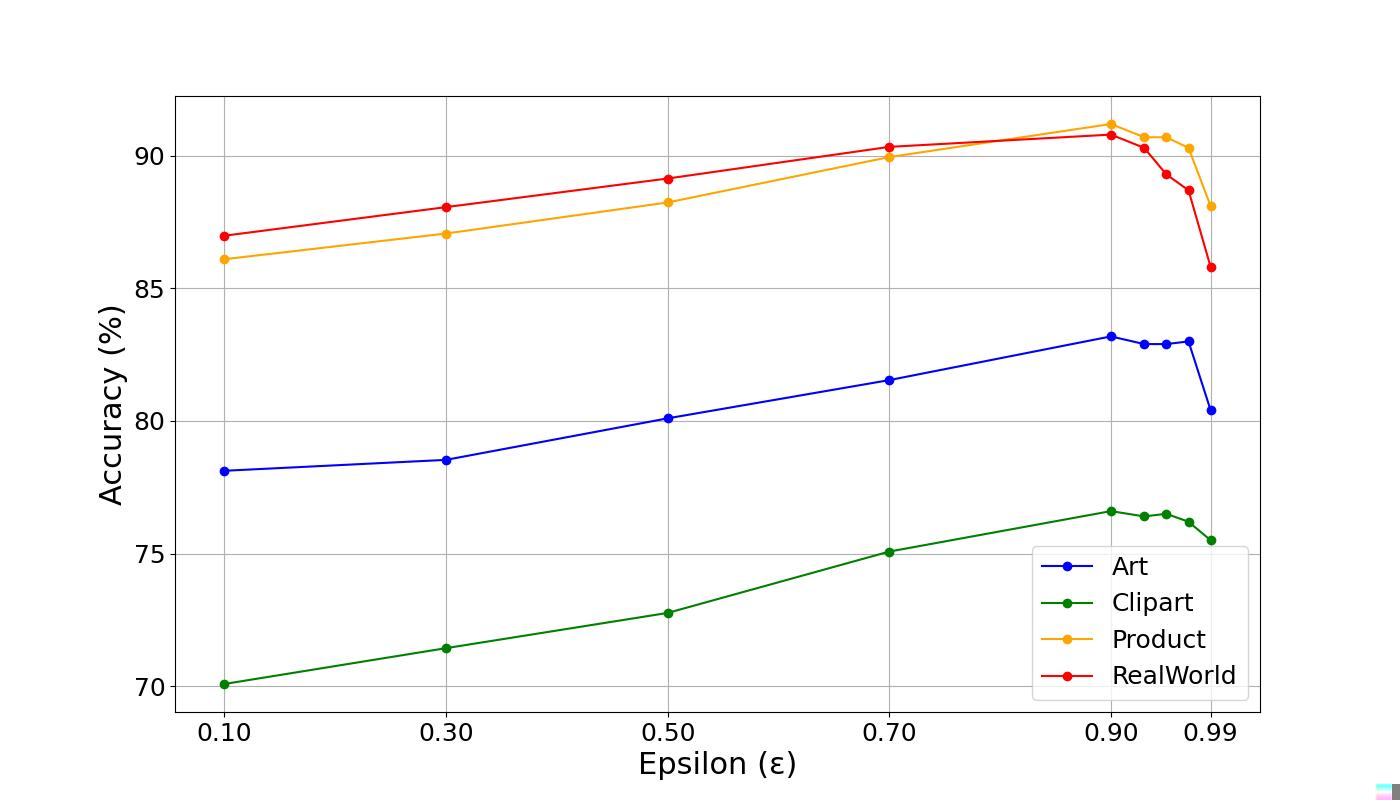}
    \caption{Sensitivity analysis of the proposed model. Impact of $\epsilon$ on SSCE loss.}
    \label{fig:hp}
\end{figure*}

\textbf{Effect of $\epsilon$.} Figure \ref{fig:hp} examines the effect of $\epsilon$ in Eq. \ref{eq:ssce} on performance. To this end, we experimented with various $\epsilon$ values ranging from 0.1 to 0.99. As shown in Figure \ref{fig:hp}, the model's accuracy improves as $\epsilon$ increases until $\epsilon = 0.9$ and there is a downward trend afterward. The role of $\epsilon$ is to smooth the pseudo labels, reducing their confidence while keeping the important information. Since pseudo-labels are inherently noisy, we mitigate this noise by smoothing them rather than discarding potentially informative labels, which may still contribute valuable information to the model. However, as $\epsilon$ increases, the model becomes less reliant on the exact values of $\tilde{y}_j^T$, effectively minimizing the effect of $\eta_j$, and leading to better generalization.

\textbf{Effect of different loss functions in MSPL.} We test the effects of different loss functions, including CE, soft-label CE, and SSCE, in the MSPL module. It can be seen in Table \ref{table:MSPL_oh} that SSCE, used in our method, yields the best performance. Soft-label CE (without smoothing), compared to hard-label CE (second row) shows worse results. This is because soft CE uses soft pseudo labels without any smoothing. In this case, all classes are assigned a probability, introducing more noise compared to hard-labels. However, when using SSCE, the probabilities are smoothed, which helps control the noise. Additionally, SSCE prevents over-reliance on pseudo labels, unlike hard-label CE, leading to better generalization. This explains why SSCE outperforms both hard-label CE and soft CE.

\begin{table}[t]
\caption{The impact of different loss functions on the performance of MSPL on OfficeHome.}
\label{table:MSPL_oh}
\centering
\def\arraystretch{1.2}
\resizebox{0.45\textwidth}{!}{
\begin{tabular}{l|cccc|c}
\toprule
$\mathcal{L}$ & $\rightarrow$A & $\rightarrow$C & $\rightarrow$P & $\rightarrow$R & Avg. \\ \midrule
SSCE & 83.2 & 76.6 & 91.2 & 90.8 & \textbf{85.4} \\
CE & 77.8 & 85.4 & 85.4 & 86.3 & 83.7 \\
Soft-label CE & 74.2 & 69.1 & 85.7 & 85.1 & 78.5 \\
 \bottomrule
\end{tabular}%
}
\end{table}

\textbf{SEA vs. other aggregation methods.} 
To assess the effectiveness of SEA, it evaluated against some of the aggregation methods in the same one-shot FL environment. We first averaged the source models, equivalent to setting $w$ in Eq. \ref{eq:agg_model} to $\frac{1}{M}$. Next, we applied the FedAvg model, which sets $w$ to the normalized number of samples in each domain, though this assumes domain information sharing, which violates our privacy constraint. We also tested entropy weights without scaling. The result (Table \ref{table:seba_oh}) indicates that our proposed scaled entropy attention outperforms these methods, demonstrating the effectiveness of SEA's adaptive weighting.

\begin{table}[t]
\caption{Results of SEA and FedAvg agregations on OfficeHome.}
\label{table:seba_oh}
\centering
\def\arraystretch{1.2}
\resizebox{0.5\textwidth}{!}{
\begin{tabular}{l|cccc|c}
\toprule
Aggregation Method                    & $\rightarrow$A & $\rightarrow$C & $\rightarrow$P & $\rightarrow$R & Avg. \\ \midrule
Averaging                          & 64.5           & 71.7           & 88.3           & 79.4           & 76.0 \\
FedAvg                             & 64.3           & 71.8           & 88.5           & 80.0           & 76.2 \\
Normalized Entropies               & 71.6           & 71.5           & 88.8           & 82.4           & 78.6 \\
\textbf{SEA}                       & 75.9           & 71.3           & 89.1           & 83.6           & \textbf{80.0} \\
\bottomrule
\end{tabular}%
}
\end{table}

\textbf{Effect of different backbones.}
\label{exp:backbones}
Table \ref{table:backbones} compares our proposed method with FedAvg, using various backbone architectures. The backbones considered include ViT \cite{dosovitskiy_image_2021}, Swin Transformer \cite{liu_swin_2021}, and ResNet-50 \cite{he_deep_2016}, each pre-trained on the ImageNet dataset. This comparison serves to validate the effectiveness of our SEA module for model aggregation in an FL environment and also demonstrates the MSPL module’s capacity to adapt the global model effectively to the target domain, independent of the chosen backbone. The consistent performance across these diverse architectures emphasizes the robustness of our framework and its adaptability to real-world scenarios where pre-trained models may differ.

\begin{table}[htbp]
\caption{Performance comparison of our method with different backbones on OfficeHome dataset.}
\label{table:backbones}
\centering
\def\arraystretch{1.2}
\resizebox{0.6\textwidth}{!}{
\begin{tabular}{l|c|cccc|c}
\toprule
Backbone & Method & $\rightarrow$A & $\rightarrow$C & $\rightarrow$P & $\rightarrow$R & Avg. \\ \midrule
\multirow{3}{*}{Resnet-50 \cite{he_deep_2016}} &
                                    FedAvg & 27.9 & 36.2 & 63.8 & 64.7 & 48.1 \\
                                    & SEA & 55.8 & 35.1 & 75.6 & 68.7 & 58.8 \\
                                    & SEA + MSPL & 65.0 & 44.6 & 75.6 & 77.9 & 65.8 \\ \midrule

\multirow{3}{*}{ViT \cite{dosovitskiy_image_2021}} &
                                    FedAvg & 50.8 & 53.2 & 72.2 & 76.9 & 63.3 \\
                                    & SEA & 63.5 & 53.5 & 75.9 & 79.2 & 68.0 \\
                                    & SEA + MSPL & 69.7 & 59.1 & 80.5 & 82.0 & 72.8 \\ \midrule

\multirow{3}{*}{Swin \cite{liu_swin_2021}} &
                                    FedAvg & 34.4 & 47.0 & 79.0 & 75.8 & 59.0 \\
                                    & SEA & 67.2 & 49.2 & 81.7 & 77.6 & 68.9 \\
                                    & SEA + MSPL & 73.3 & 57.4 & 83.7 & 83.2 & 74.4 \\ \midrule

\multirow{3}{*}{DINOv2 \cite{oquab_dinov2_2024}} &
                                    FedAvg & 64.3 & 71.8 & 88.5 & 80.0 & 76.2 \\
                                    & SEA & 75.9 & 71.3 & 89.1 & 83.6 & 80.0 \\
                                    & SEA + MSPL & \textbf{83.2} & \textbf{76.6} & \textbf{91.2} & \textbf{90.8} & \textbf{85.4} \\
                                                        
\bottomrule
\end{tabular}%
}
\end{table}

\begin{figure*}[t]
\centering
\includegraphics[width=\textwidth]{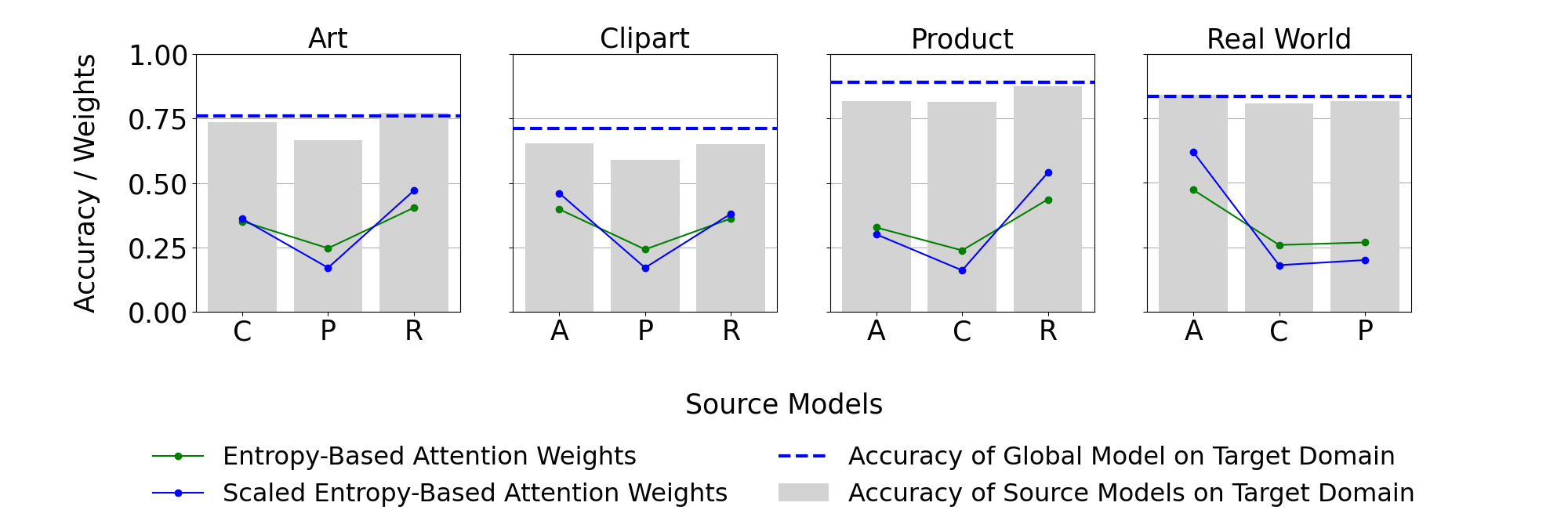}
\caption{Accuracy of source models for each OfficeHome target domain.}
\label{fig:enatt}
\end{figure*}

\textbf{Effect of scaling in SEA.} Here, we study the impact of scaling on SEA. The results (Figure \ref{fig:enatt}) show that models with higher accuracy are assigned higher entropy weights. By applying scaling, higher-confident models with better performance receive greater attention in the aggregation process. This approach allows stronger models to have a more substantial influence on the aggregation while reducing the risk of overemphasizing any single model and thereby enhancing the overall performance of global model on unlabeled target domain and robustness.

\begin{figure*}[t]
    \centering
    \begin{subfigure}{0.24\textwidth}
        \centering
        \includegraphics[width=\linewidth]{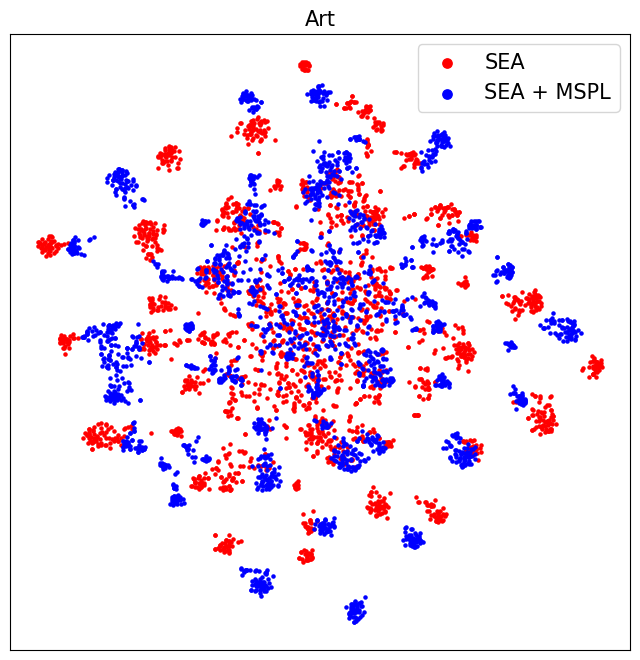}
        \caption{Art}
        \label{fig:tsne_art}
    \end{subfigure}
    \begin{subfigure}{0.24\textwidth}
        \centering
        \includegraphics[width=\linewidth]{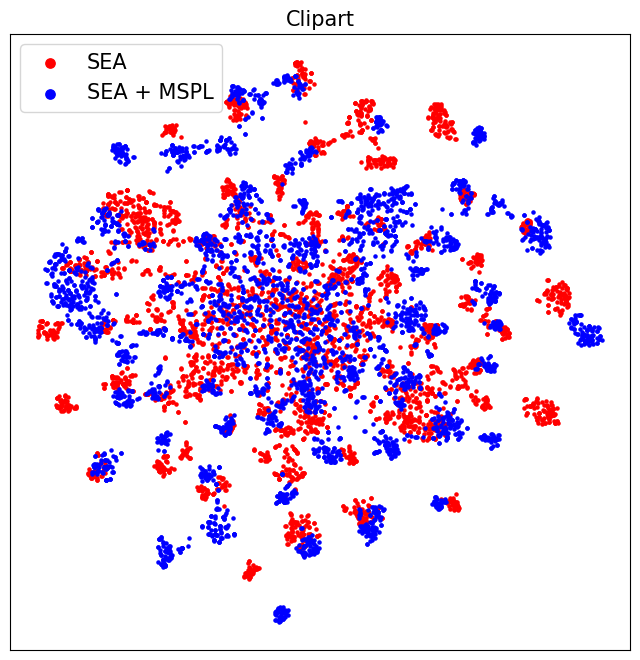}
        \caption{Clipart}
        \label{fig:tsne_clipart}
    \end{subfigure}
    \begin{subfigure}{0.24\textwidth}
        \centering
        \includegraphics[width=\linewidth]{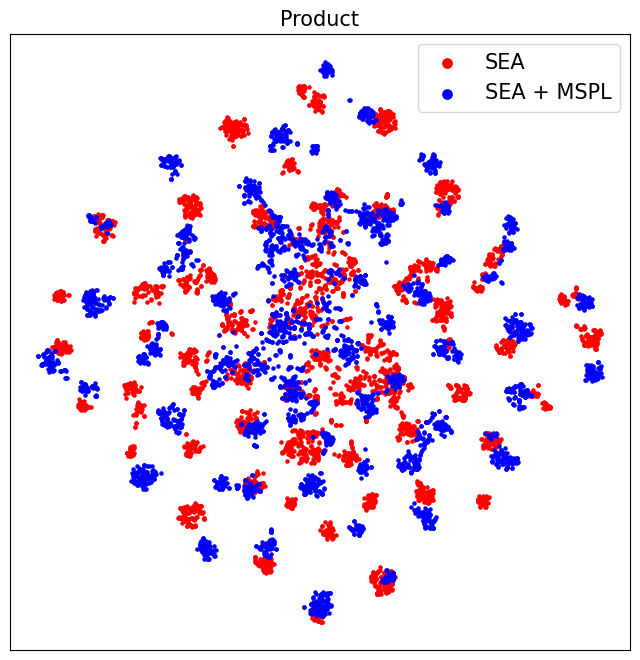}
        \caption{Product}
        \label{fig:tsne_product}
    \end{subfigure}
    \begin{subfigure}{0.24\textwidth}
        \centering
        \includegraphics[width=\linewidth]{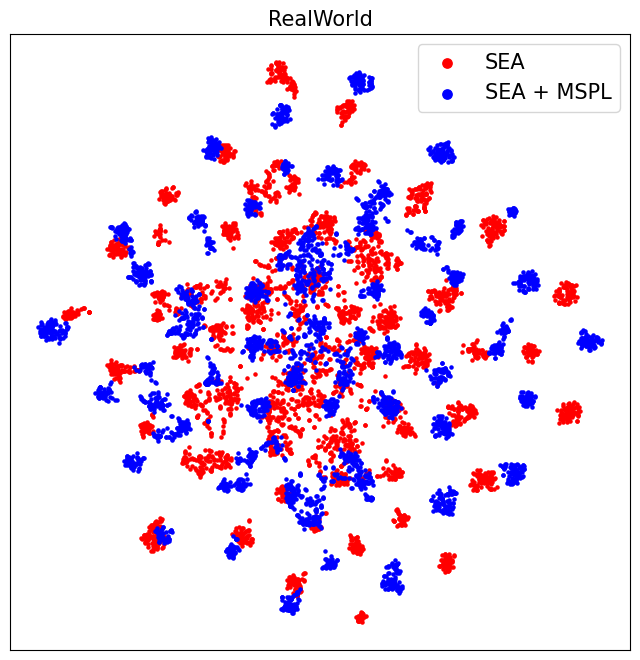}
        \caption{RealWorld}
        \label{fig:tsne_realworld}
    \end{subfigure}
    \caption{t-SNE visualizations of target domain feature distributions for both SEA and SEA + MSPL.}
    \label{fig:tsne}
\end{figure*}

\textbf{t-SNE analysis of features.}
Figure \ref{fig:tsne} presents the t-SNE visualizations of the target domain feature distributions for SEA and SEA + MSPL. The t-SNE plot for SEA shows well-formed clusters, indicating good feature alignment and adaptation. However, the plot for SEA + MSPL shows even better-separated clusters, demonstrating improved feature differentiation and further reducing overlap between classes. This indicates that the MSPL module enhances the alignment of target domain features, providing additional refinement and improving adaptation performance compared to SEA alone.

\section{Conclusion}
\label{sec:conclusion}
In this work, we present a novel and efficient solution for FUDA that tackles critical challenges such as domain shifts, communication overhead, and computational efficiency. Our one-shot FL setup significantly reduces communication costs by limiting interactions between clients and the server to a single round. The model is also computationally efficient, as each client shares only the small bottleneck and head parameters, minimizing the number of trainable parameters. We introduce SEA to balance model contributions by assigning attention based on scaled prediction entropy, ensuring that stronger models are prioritized without ignoring weaker ones. Additionally, we introduce MSPL to enhance target adaptation by generating pseudo labels and refining them with SSCE to handle noise. Extensive evaluations across four benchmarks show that our method outperforms SOTA approaches while maintaining high efficiency and privacy. Ablation studies further validate the effectiveness of each component, making this approach suitable for real-world, resource-constrained FL applications. 
In future work, we aim to apply attention and weighting into pseudo-label generation to enhance accuracy and further reduce noise. Additionally, we plan to extend the proposed framework to other downstream tasks where domain adaptation is essential, such as object detection.


\begin{thebibliography}{10}

\bibitem{mcmahan_communication-efficient_2017}
Brendan McMahan, Eider Moore, Daniel Ramage, Seth Hampson, and Blaise Aguera~y Arcas.
\newblock Communication-{Efficient} {Learning} of {Deep} {Networks} from {Decentralized} {Data}.
\newblock In Aarti Singh and Jerry Zhu, editors, {\em Proceedings of the 20th {International} {Conference} on {Artificial} {Intelligence} and {Statistics}}, volume~54 of {\em Proceedings of {Machine} {Learning} {Research}}, pages 1273--1282. PMLR, April 2017.

\bibitem{zhang_survey_2021}
Chen Zhang, Yu~Xie, Hang Bai, Bin Yu, Weihong Li, and Yuan Gao.
\newblock A survey on federated learning.
\newblock {\em Knowledge-Based Systems}, 216:106775, March 2021.

\bibitem{shenaj_federated_2023}
Donald Shenaj, Giulia Rizzoli, and Pietro Zanuttigh.
\newblock Federated {Learning} in {Computer} {Vision}.
\newblock {\em IEEE Access}, 11:94863--94884, 2023.

\bibitem{wen_survey_2023}
Jie Wen, Zhixia Zhang, Yang Lan, Zhihua Cui, Jianghui Cai, and Wensheng Zhang.
\newblock A survey on federated learning: challenges and applications.
\newblock {\em International Journal of Machine Learning and Cybernetics}, 14(2):513--535, February 2023.

\bibitem{beitollahi_parametric_2024}
Mahdi Beitollahi, Alex Bie, Sobhan Hemati, Leo~Maxime Brunswic, Xu~Li, Xi~Chen, and Guojun Zhang.
\newblock Parametric {Feature} {Transfer}: {One}-shot {Federated} {Learning} with {Foundation} {Models}, February 2024.
\newblock arXiv:2402.01862 [cs].

\bibitem{feng_kd3a_2021}
Haozhe Feng, Zhaoyang You, Minghao Chen, Tianye Zhang, Minfeng Zhu, Fei Wu, Chao Wu, and Wei Chen.
\newblock {KD3A}: {Unsupervised} {Multi}-{Source} {Decentralized} {Domain} {Adaptation} via {Knowledge} {Distillation}.
\newblock In {\em Proceedings of the 38th {International} {Conference} on {Machine} {Learning}}, pages 3274--3283. PMLR, July 2021.
\newblock ISSN: 2640-3498.

\bibitem{yi_model-contrastive_2023}
Chang'an Yi, Haotian Chen, Yonghui Xu, and Yifan Zhang.
\newblock Model-{Contrastive} {Federated} {Domain} {Adaptation}, May 2023.
\newblock arXiv:2305.10432 version: 1.

\bibitem{li_federated_2024}
Ying Li, Xingwei Wang, Rongfei Zeng, Praveen~Kumar Donta, Ilir Murturi, Min Huang, and Schahram Dustdar.
\newblock Federated {Domain} {Generalization}: {A} {Survey}, March 2024.
\newblock arXiv:2306.01334.

\bibitem{wang_deep_2018}
Mei Wang and Weihong Deng.
\newblock Deep visual domain adaptation: {A} survey.
\newblock {\em Neurocomputing}, 312:135--153, October 2018.

\bibitem{guha_one-shot_2019}
Neel Guha, Ameet Talwalkar, and Virginia Smith.
\newblock One-{Shot} {Federated} {Learning}, March 2019.
\newblock arXiv:1902.11175 [cs, stat].

\bibitem{su_one-shot_2023}
Shangchao Su, Bin Li, and Xiangyang Xue.
\newblock One-shot {Federated} {Learning} without server-side training.
\newblock {\em Neural Networks}, 164:203--215, July 2023.

\bibitem{sun_survey_2015}
Shiliang Sun, Honglei Shi, and Yuanbin Wu.
\newblock A survey of multi-source domain adaptation.
\newblock {\em Information Fusion}, 24:84--92, July 2015.

\bibitem{li_multi-source_2022}
Keqiuyin Li, Jie Lu, Hua Zuo, and Guangquan Zhang.
\newblock Multi-{Source} {Contribution} {Learning} for {Domain} {Adaptation}.
\newblock {\em IEEE Transactions on Neural Networks and Learning Systems}, 33(10):5293--5307, October 2022.

\bibitem{yin_universal_2022}
Yueming Yin, Zhen Yang, Haifeng Hu, and Xiaofu Wu.
\newblock Universal multi-{Source} domain adaptation for image classification.
\newblock {\em Pattern Recognition}, 121:108238, January 2022.

\bibitem{wei_multi-source_2024}
Yikang Wei and Yahong Han.
\newblock Multi-{Source} {Collaborative} {Gradient} {Discrepancy} {Minimization} for {Federated} {Domain} {Generalization}.
\newblock {\em Proceedings of the AAAI Conference on Artificial Intelligence}, 38(14):15805--15813, March 2024.
\newblock Number: 14.

\bibitem{kang_communicational_2022}
Hua Kang, Zhiyang Li, and Qian Zhang.
\newblock Communicational and {Computational} {Efficient} {Federated} {Domain} {Adaptation}.
\newblock {\em IEEE Transactions on Parallel and Distributed Systems}, 33(12):3678--3689, December 2022.

\bibitem{peng_federated_2019}
Xingchao Peng, Zijun Huang, Yizhe Zhu, and Kate Saenko.
\newblock Federated {Adversarial} {Domain} {Adaptation}.
\newblock In {\em International {Conference} on {Learning} {Representations}}, December 2019.

\bibitem{pourpanah_federated_2025}
Farhad Pourpanah, Mahdiyar Molahasani, Milad Soltany, Michael Greenspan, and Ali Etemad.
\newblock Federated {Unsupervised} {Domain} {Generalization} using {Global} and {Local} {Alignment} of {Gradients}, January 2025.
\newblock arXiv:2405.16304 [cs].

\bibitem{liu_projected_2021}
Junxu Liu, Jian Lou, Li~Xiong, Jinfei Liu, and Xiaofeng Meng.
\newblock Projected federated averaging with heterogeneous differential privacy.
\newblock {\em Proceedings of the VLDB Endowment}, 15(4):828--840, December 2021.

\bibitem{li_joint_2022}
Yifei Li, Yijia Guo, Mamoun Alazab, Shengbo Chen, Cong Shen, and Keping Yu.
\newblock Joint {Optimal} {Quantization} and {Aggregation} of {Federated} {Learning} {Scheme} in {VANETs}.
\newblock {\em IEEE Transactions on Intelligent Transportation Systems}, 23(10):19852--19863, October 2022.

\bibitem{qi_model_2024}
Pian Qi, Diletta Chiaro, Antonella Guzzo, Michele Ianni, Giancarlo Fortino, and Francesco Piccialli.
\newblock Model aggregation techniques in federated learning: {A} comprehensive survey.
\newblock {\em Future Generation Computer Systems}, 150:272--293, January 2024.

\bibitem{saito_asymmetric_2017}
Kuniaki Saito, Yoshitaka Ushiku, and Tatsuya Harada.
\newblock Asymmetric {Tri}-training for {Unsupervised} {Domain} {Adaptation}.
\newblock In {\em Proceedings of the 34th {International} {Conference} on {Machine} {Learning}}, pages 2988--2997. PMLR, July 2017.
\newblock ISSN: 2640-3498.

\bibitem{zou_unsupervised_2018}
Yang Zou, Zhiding Yu, B.~V. K.~Vijaya Kumar, and Jinsong Wang.
\newblock Unsupervised {Domain} {Adaptation} for {Semantic} {Segmentation} via {Class}-{Balanced} {Self}-{Training}.
\newblock In {\em Proceedings of the {European} {Conference} on {Computer} {Vision} ({ECCV})}, pages 289--305, 2018.

\bibitem{szegedy_rethinking_2016}
Christian Szegedy, Vincent Vanhoucke, Sergey Ioffe, Jon Shlens, and Zbigniew Wojna.
\newblock Rethinking the {Inception} {Architecture} for {Computer} {Vision}.
\newblock In {\em 2016 {IEEE} {Conference} on {Computer} {Vision} and {Pattern} {Recognition} ({CVPR})}, pages 2818--2826, June 2016.
\newblock ISSN: 1063-6919.

\bibitem{soltany_federated_2025}
Milad Soltany, Farhad Pourpanah, Mahdiyar Molahasani, Michael Greenspan, and Ali Etemad.
\newblock Federated {Domain} {Generalization} with {Label} {Smoothing} and {Balanced} {Decentralized} {Training}, February 2025.
\newblock arXiv:2412.11408 [cs].

\bibitem{venkateswara_deep_2017}
Hemanth Venkateswara, Jose Eusebio, Shayok Chakraborty, and Sethuraman Panchanathan.
\newblock Deep {Hashing} {Network} for {Unsupervised} {Domain} {Adaptation}.
\newblock In {\em Proceedings of the {IEEE} {Conference} on {Computer} {Vision} and {Pattern} {Recognition}}, pages 5018--5027, 2017.

\bibitem{hutchison_adapting_2010}
David Hutchison, Takeo Kanade, Josef Kittler, Jon~M. Kleinberg, Friedemann Mattern, John~C. Mitchell, Moni Naor, Oscar Nierstrasz, C.~Pandu~Rangan, Bernhard Steffen, Madhu Sudan, Demetri Terzopoulos, Doug Tygar, Moshe~Y. Vardi, Gerhard Weikum, Kate Saenko, Brian Kulis, Mario Fritz, and Trevor Darrell.
\newblock Adapting {Visual} {Category} {Models} to {New} {Domains}.
\newblock In Kostas Daniilidis, Petros Maragos, and Nikos Paragios, editors, {\em Computer {Vision} - {ECCV} 2010}, volume 6314, pages 213--226. Springer Berlin Heidelberg, Berlin, Heidelberg, 2010.
\newblock Series Title: Lecture Notes in Computer Science.

\bibitem{gong_geodesic_2012}
Boqing Gong, Yuan Shi, Fei Sha, and Kristen Grauman.
\newblock Geodesic flow kernel for unsupervised domain adaptation.
\newblock In {\em 2012 {IEEE} {Conference} on {Computer} {Vision} and {Pattern} {Recognition}}, pages 2066--2073, June 2012.
\newblock ISSN: 1063-6919.

\bibitem{peng_moment_2019}
Xingchao Peng, Qinxun Bai, Xide Xia, Zijun Huang, Kate Saenko, and Bo~Wang.
\newblock Moment {Matching} for {Multi}-{Source} {Domain} {Adaptation}.
\newblock In {\em Proceedings of the {IEEE}/{CVF} {International} {Conference} on {Computer} {Vision} ({ICCV})}, October 2019.

\bibitem{yurochkin_bayesian_2019}
Mikhail Yurochkin, Mayank Agarwal, Soumya Ghosh, Kristjan Greenewald, Nghia Hoang, and Yasaman Khazaeni.
\newblock Bayesian {Nonparametric} {Federated} {Learning} of {Neural} {Networks}.
\newblock In {\em Proceedings of the 36th {International} {Conference} on {Machine} {Learning}}, pages 7252--7261. PMLR, May 2019.
\newblock ISSN: 2640-3498.

\bibitem{liang_we_2020}
Jian Liang, Dapeng Hu, and Jiashi Feng.
\newblock Do {We} {Really} {Need} to {Access} the {Source} {Data}? {Source} {Hypothesis} {Transfer} for {Unsupervised} {Domain} {Adaptation}.
\newblock In {\em Proceedings of the 37th {International} {Conference} on {Machine} {Learning}}, pages 6028--6039. PMLR, November 2020.
\newblock ISSN: 2640-3498.

\bibitem{zuo_attention-based_2021}
Yukun Zuo, Hantao Yao, and Changsheng Xu.
\newblock Attention-{Based} {Multi}-{Source} {Domain} {Adaptation}.
\newblock {\em IEEE Transactions on Image Processing}, 30:3793 -- 3803, March 2021.

\bibitem{venkat_your_2020}
Naveen Venkat, Jogendra~Nath Kundu, Durgesh Singh, Ambareesh Revanur, and Venkatesh~Babu R.
\newblock Your {Classifier} can {Secretly} {Suffice} {Multi}-{Source} {Domain} {Adaptation}.
\newblock In {\em Advances in {Neural} {Information} {Processing} {Systems}}, volume~33, pages 4647--4659. Curran Associates, Inc., 2020.

\bibitem{zhou_cycle_2024}
Chaoyang Zhou, Zengmao Wang, Bo~Du, and Yong Luo.
\newblock Cycle {Self}-{Refinement} for {Multi}-{Source} {Domain} {Adaptation}.
\newblock {\em Proceedings of the AAAI Conference on Artificial Intelligence}, 38(15):17096--17104, March 2024.
\newblock Number: 15.

\bibitem{ahmed_unsupervised_2021}
Sk~Miraj Ahmed, Dripta~S. Raychaudhuri, Sujoy Paul, Samet Oymak, and Amit~K. Roy-Chowdhury.
\newblock Unsupervised {Multi}-{Source} {Domain} {Adaptation} {Without} {Access} to {Source} {Data}.
\newblock In {\em Proceedings of the {IEEE}/{CVF} {Conference} on {Computer} {Vision} and {Pattern} {Recognition} ({CVPR})}, pages 10103--10112, June 2021.

\bibitem{pei_evidential_2024}
Jiangbo Pei, Aidong Men, Yang Liu, Xiahai Zhuang, and Qingchao Chen.
\newblock Evidential {Multi}-{Source}-{Free} {Unsupervised} {Domain} {Adaptation}.
\newblock {\em IEEE Transactions on Pattern Analysis and Machine Intelligence}, 46(8):5288--5305, August 2024.

\bibitem{li_agile_2024}
Xinyao Li, Jingjing Li, Fengling Li, Lei Zhu, and Ke~Lu.
\newblock Agile {Multi}-{Source}-{Free} {Domain} {Adaptation}.
\newblock {\em Proceedings of the AAAI Conference on Artificial Intelligence}, 38(12):13673--13681, March 2024.
\newblock Number: 12.

\bibitem{wu_collaborative_2021}
Guile Wu and Shaogang Gong.
\newblock Collaborative {Optimization} and {Aggregation} for {Decentralized} {Domain} {Generalization} and {Adaptation}.
\newblock In {\em Proceedings of the {IEEE}/{CVF} {International} {Conference} on {Computer} {Vision}}, pages 6484--6493, 2021.

\bibitem{niu_mckd_2023}
Ziwei Niu, Hongyi Wang, Hao Sun, Shuyi Ouyang, Yen-wei Chen, and Lanfen Lin.
\newblock {MCKD}: {Mutually} {Collaborative} {Knowledge} {Distillation} {For} {Federated} {Domain} {Adaptation} {And} {Generalization}.
\newblock In {\em {ICASSP} 2023 - 2023 {IEEE} {International} {Conference} on {Acoustics}, {Speech} and {Signal} {Processing} ({ICASSP})}, pages 1--5, June 2023.
\newblock ISSN: 2379-190X.

\bibitem{li_federated_2024-1}
Keqiuyin Li, Jie Lu, Hua Zuo, and Guangquan Zhang.
\newblock Federated {Fuzzy} {Transfer} {Learning} {With} {Domain} and {Category} {Shifts}.
\newblock {\em IEEE Transactions on Fuzzy Systems}, pages 1--12, 2024.
\newblock Conference Name: IEEE Transactions on Fuzzy Systems.

\bibitem{liu_ufda_2024}
Xinhui Liu, Zhenghao Chen, Luping Zhou, Dong Xu, Wei Xi, Gairui Bai, Yihan Zhao, and Jizhong Zhao.
\newblock {UFDA}: {Universal} {Federated} {Domain} {Adaptation} with {Practical} {Assumptions}.
\newblock {\em Proceedings of the AAAI Conference on Artificial Intelligence}, 38(12):14026--14034, March 2024.
\newblock Number: 12.

\bibitem{li_pseudo_2023}
Yundong Li, Longxia Guo, and Yizheng Ge.
\newblock Pseudo {Labels} for {Unsupervised} {Domain} {Adaptation}: {A} {Review}.
\newblock {\em Electronics}, 12(15):3325, August 2023.

\bibitem{wang_unsupervised_2020}
Qian Wang and Toby Breckon.
\newblock Unsupervised {Domain} {Adaptation} via {Structured} {Prediction} {Based} {Selective} {Pseudo}-{Labeling}.
\newblock {\em Proceedings of the AAAI Conference on Artificial Intelligence}, 34(04):6243--6250, April 2020.
\newblock Number: 04.

\bibitem{litrico_guiding_2023}
Mattia Litrico, Alessio Del~Bue, and Pietro Morerio.
\newblock Guiding {Pseudo}-{Labels} {With} {Uncertainty} {Estimation} for {Source}-{Free} {Unsupervised} {Domain} {Adaptation}.
\newblock In {\em Proceedings of the {IEEE}/{CVF} {Conference} on {Computer} {Vision} and {Pattern} {Recognition}}, pages 7640--7650, 2023.

\bibitem{matsuzaki_multi-source_2023}
Shigemichi Matsuzaki, Hiroaki Masuzawa, and Jun Miura.
\newblock Multi-{Source} {Soft} {Pseudo}-{Label} {Learning} with {Domain} {Similarity}-based {Weighting} for {Semantic} {Segmentation}.
\newblock In {\em 2023 {IEEE}/{RSJ} {International} {Conference} on {Intelligent} {Robots} and {Systems} ({IROS})}, pages 5852--5857, October 2023.
\newblock ISSN: 2153-0866.

\bibitem{zheng_rectifying_2021}
Zhedong Zheng and Yi~Yang.
\newblock Rectifying {Pseudo} {Label} {Learning} via {Uncertainty} {Estimation} for {Domain} {Adaptive} {Semantic} {Segmentation}.
\newblock {\em International Journal of Computer Vision}, 129(4):1106--1120, April 2021.

\bibitem{oquab_dinov2_2024}
Maxime Oquab, Timothée Darcet, Théo Moutakanni, Huy Vo, Marc Szafraniec, Vasil Khalidov, Pierre Fernandez, Daniel Haziza, Francisco Massa, Alaaeldin El-Nouby, Mahmoud Assran, Nicolas Ballas, Wojciech Galuba, Russell Howes, Po-Yao Huang, Shang-Wen Li, Ishan Misra, Michael Rabbat, Vasu Sharma, Gabriel Synnaeve, Hu~Xu, Hervé Jegou, Julien Mairal, Patrick Labatut, Armand Joulin, and Piotr Bojanowski.
\newblock {DINOv2}: {Learning} {Robust} {Visual} {Features} without {Supervision}, February 2024.
\newblock arXiv:2304.07193 [cs].

\bibitem{dosovitskiy_image_2021}
Alexey Dosovitskiy, Lucas Beyer, Alexander Kolesnikov, Dirk Weissenborn, Xiaohua Zhai, Thomas Unterthiner, Mostafa Dehghani, Matthias Minderer, Georg Heigold, Sylvain Gelly, Jakob Uszkoreit, and Neil Houlsby.
\newblock An {Image} is {Worth} 16x16 {Words}: {Transformers} for {Image} {Recognition} at {Scale}.
\newblock {\em ICLR}, 2021.

\bibitem{abedi_euda_2024}
Ali Abedi, Q.~M.~Jonathan Wu, Ning Zhang, and Farhad Pourpanah.
\newblock {EUDA}: {An} {Efficient} {Unsupervised} {Domain} {Adaptation} via {Self}-{Supervised} {Vision} {Transformer}, July 2024.
\newblock arXiv:2407.21311 [cs].

\bibitem{louizos_multiplicative_2017}
Christos Louizos and Max Welling.
\newblock Multiplicative {Normalizing} {Flows} for {Variational} {Bayesian} {Neural} {Networks}.
\newblock In Doina Precup and Yee~Whye Teh, editors, {\em Proceedings of the 34th {International} {Conference} on {Machine} {Learning}}, volume~70 of {\em Proceedings of {Machine} {Learning} {Research}}, pages 2218--2227. PMLR, August 2017.

\bibitem{sensoy_evidential_2018}
Murat Sensoy, Lance Kaplan, and Melih Kandemir.
\newblock Evidential {Deep} {Learning} to {Quantify} {Classification} {Uncertainty}.
\newblock In {\em Advances in {Neural} {Information} {Processing} {Systems}}, volume~31. Curran Associates, Inc., 2018.

\bibitem{mansour_domain_2008}
Yishay Mansour, Mehryar Mohri, and Afshin Rostamizadeh.
\newblock Domain {Adaptation} with {Multiple} {Sources}.
\newblock In D.~Koller, D.~Schuurmans, Y.~Bengio, and L.~Bottou, editors, {\em Advances in {Neural} {Information} {Processing} {Systems}}, volume~21. Curran Associates, Inc., 2008.

\bibitem{he_deep_2016}
Kaiming He, Xiangyu Zhang, Shaoqing Ren, and Jian Sun.
\newblock Deep {Residual} {Learning} for {Image} {Recognition}.
\newblock In {\em 2016 {IEEE} {Conference} on {Computer} {Vision} and {Pattern} {Recognition} ({CVPR})}, pages 770--778, Las Vegas, NV, USA, June 2016. IEEE.

\bibitem{liu_swin_2021}
Ze~Liu, Yutong Lin, Yue Cao, Han Hu, Yixuan Wei, Zheng Zhang, Stephen Lin, and Baining Guo.
\newblock Swin {Transformer}: {Hierarchical} {Vision} {Transformer} {Using} {Shifted} {Windows}.
\newblock In {\em Proceedings of the {IEEE}/{CVF} {International} {Conference} on {Computer} {Vision} ({ICCV})}, pages 10012--10022, October 2021.

\end{thebibliography}

\appendix

\section*{Supplementary Materials}

\section{Proofs}
\label{sec:proofs}

\subsection{Proof Sketch for Theorem \ref{theorem_1}}
\label{proof:theorem_1}
Multi-source domain adaptation theory typically provides a bound of the form:
\begin{equation}
R_T(f_g) \le \sum_{i=1}^{M} w_i R_T(f_i) + \Phi.
\end{equation}

By substituting $R_T(f_i) \approx k\,\bar{H}_i$ (Assumption \ref{assumption_1}), the bound becomes:
\begin{equation}
R_T(f_g) \lesssim k\sum_{i=1}^{M} w_i\, \bar{H}_i + \Phi.
\end{equation}

\hfill \(\square\)

\subsection{Proof Sketch for Lemma \ref{lemma_2}}
\label{proof:lemma_2}
\noindent\textbf{Proof.}  
Given that entropy values are closely clustered, we have:

\begin{equation}
|\bar{H}_{i} - \bar{H}_{j}| \approx 0, \quad \forall i, j.
\end{equation}

Since the unnormalized weights are defined as:

\begin{equation}
w'_i = \frac{1}{\bar{H}_i},
\end{equation}
taking the difference between any two weights gives:

\begin{equation}
\left| w'_i - w'_j \right| = \left| \frac{1}{\bar{H}_i} - \frac{1}{\bar{H}_j} \right|.
\end{equation}

Using the first-order approximation for small perturbations,

\begin{equation}
\frac{1}{\bar{H}_i} - \frac{1}{\bar{H}_j} \approx -\frac{\bar{H}_j - \bar{H}_i}{\bar{H}_i \bar{H}_j},
\end{equation}
which approaches zero as \( \bar{H}_i \approx \bar{H}_j \). Thus, we obtain:

\begin{equation}
|w'_i - w'_j| \approx 0.
\end{equation}

Normalizing these weights,

\begin{equation}
w_i = \frac{w'_i}{\sum_{j=1}^{M} w'_j},
\end{equation}
and since \( w'_i \approx w'_j \), the sum in the denominator satisfies:

\begin{equation}
\sum_{j=1}^{M} w'_j \approx M \mu_{w'},
\end{equation}
where the mean weight is:

\begin{equation}
\mu_{w'} = \frac{1}{M} \sum_{j=1}^{M} w'_j.
\end{equation}

Thus, the normalized weights satisfy:

\begin{equation}
w_i \approx \frac{w'_i}{M \mu_{w'}} \approx \frac{1}{M}.
\end{equation}

Since this holds for all \( i \), the weights converge toward their mean, resulting in a nearly uniform distribution:

\begin{equation}
w_i \approx w_j \approx \frac{1}{M}, \quad \forall i, j.
\end{equation}

\hfill \(\square\)

\subsection{Proof Sketch for Lemma \ref{lemma_ce}}
\label{proof:lemma_ce}
The cross-entropy loss for a classifier \( f_\theta \) trained on pseudo labels is:

\begin{equation}
    \mathcal{L}_{\text{CE}} = - \sum_{i=1}^{C} \tilde{y}_i \log f_\theta(x)_i.
\end{equation}

For one-hot labels \( y \), this simplifies to:

\begin{equation}
    \mathcal{L}_{\text{CE}} = - \log f_\theta(x)_k,
\end{equation}

where \( k \) is the ground-truth class. This forces the model to maximize \( f_\theta(x)_k \to 1 \), causing it to overfit to training data by assigning zero probability to all incorrect classes. \hfill \(\square\)

\subsection{Proof Sketch for Theorem \ref{theorem_ls}}
\label{proof:theorem_ls}
Substituting \( \tilde{y} = y + \eta \) into the label smoothing equation:

\begin{equation}
    \tilde{y}' = (1 - \epsilon)(y + \eta) + \frac{\epsilon}{C} = (1 - \epsilon)y + (1 - \epsilon)\eta + \frac{\epsilon}{C}.
\end{equation}

Since \(\eta\) represents the pseudo-label noise, multiplying it by \((1 - \epsilon)\) reduces its magnitude, thereby limiting the influence of incorrect pseudo labels. The SSCE loss is then computed using equation \ref{eq:ssce}. By controlling \(\epsilon\), SSCE smooths the pseudo labels, reducing overconfidence while preserving useful knowledge. When \(\epsilon \to 1\), the labels approach a uniform distribution, preventing reliance on uncertain predictions, whereas for \(\epsilon \to 0\), SSCE recovers standard cross-entropy. \hfill \(\square\)

\section{More intuition behind SEA}
\label{sec:sup_seba}
The entropy of a model's predictions indicates its uncertainty \cite{louizos_multiplicative_2017, sensoy_evidential_2018}. Lower entropy reflects more certain predictions, making the model more reliable on the target domain, while higher entropy indicates less confidence. Therefore, each source model is weighted in the final aggregation based on its entropy, with lower-entropy (higher-certainty) models contributing more.

\begin{figure}[htbp]
\centering
\includegraphics[width=0.45\textwidth]{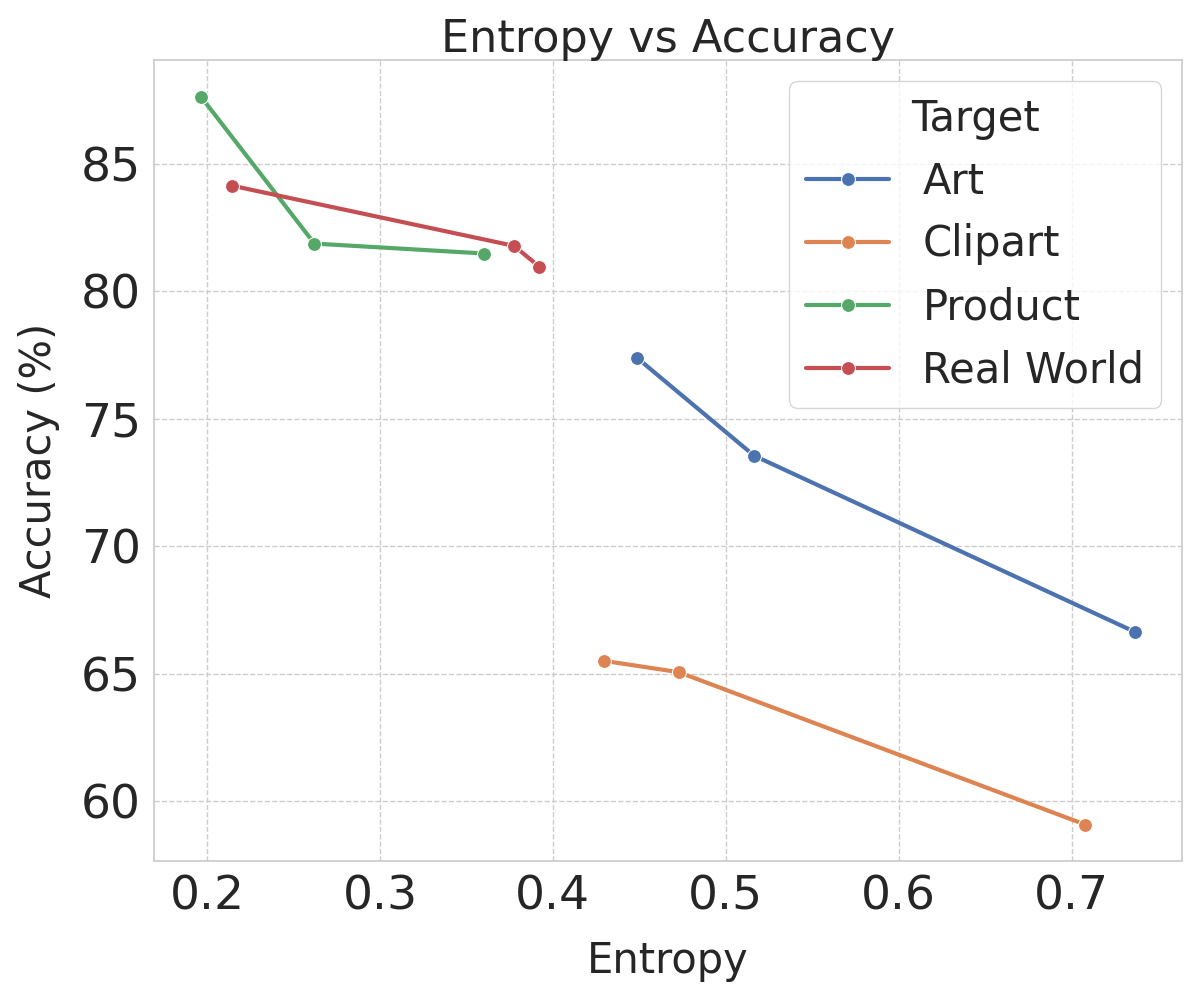}
\caption{Relationship between entropy and accuracy of source models across target domains in OfficeHome.}
\label{fig:envsacc}
\end{figure}

Low entropy signifies focused predictions on a single class (high confidence), whereas high entropy suggests uncertainty with predictions distributed across multiple classes, where entropy $H(p(x))$ is directly proportional to uncertainty. Accuracy, which reflects the frequency of correct predictions, usually is inversely related to entropy: a highly accurate model confidently assigns a high probability to the correct class, reducing entropy. For a dataset, the model's expected accuracy $A$ is defined as the average of correct predictions:

\begin{equation}
    A = \frac{1}{N} \sum_{j=1}^{N} \mathds{1}[\hat{y}_j = y_j],
\end{equation}

where $N$ is the total number of samples, $\hat{y}_j$ is the predicted label for sample $j$, $y_j$ is the true label, and $\mathds{1}[\hat{y}_j = y_j]$ is 1 if the prediction matches the true label, and 0 otherwise. When a model is accurate, $p(x)$ for the correct class approaches 1, leading to lower entropy:

\begin{equation}
    H(p(x)) \approx 0 ~\text{when}~ p(x)_\text{correct~class} \rightarrow 1.
\end{equation}

Thus, accuracy is inversely related to entropy, $A \propto \frac{1}{H(p(x))}$. 

Figure \ref{fig:envsacc} illustrates this relationship, highlighting source model performance across various target domains.

\end{document}